# Comprehensive literature survey on deep learning used in image memorability prediction and modification

Ananya Sadana[1], Nikita Thakur[2], Nikita Poria[3], Astika Anand[4] and Seeja K.R.[5]

Department of Computer Science and Engineering, Indira Gandhi Delhi Technical University for Women, Delhi, India
ananya097btcse19@igdtuw.ac.in[1], nikita112btcse19@igdtuw.ac.in[2], nikita090btcse19@igdtuw.ac.in[3], astika086btcse19@igdtuw.ac.in[4], seeja@igdtuw.ac.in[5]

**Abstract:** As humans, we can remember certain visuals in great detail, and sometimes even after viewing them once. What is even more interesting is that humans tend to remember and forget the same things, suggesting that there might be some general internal characteristics of an image that make it easier for the brain to encode and discard certain types of information. Research suggests that some pictures tend to be memorized more than others. The ability of an image to be remembered by different viewers is one of its intrinsic properties. In visualization and photography, creating memorable images is a difficult task. Hence, to solve the problem, various techniques predict visual memorability and manipulate images' memorability. We present a comprehensive literature survey to assess the deep learning techniques used to predict and modify memorability. In particular, we analyze the use of Convolutional Neural Networks, Recurrent Neural Networks, and Generative Adversarial Networks for image memorability prediction and modification.

## 1 Introduction

Every day we are exposed to many images, only a few of which are remembered, while most of them we tend to forget. Though the human cognitive system has an enormous storage capacity [1,2], it may only be able to store some images as detailed as they are. Few images are remembered in great detail, even fewer in minor details, and the remainder is quickly forgotten [3]. Natural scenery photos, for example, are less likely to be remembered than images of animals, vehicles, and people [4]. According to previous research, images are consistently memorable to different viewers [5] and some images have better memorability than others. They also showed that memorability is an intrinsic and measurable property of an image. When we discuss memorability as a measurable property, the question of an artificial system successfully predicting the image memorability score comes along.

Previous works done in the domain of image memorability can be grouped into three categories - understanding features that affect image memorability, Prediction of images' memorability scores, and modifying images' memorability. Memorability was initially calculated as a probabilistic function through various experiments conducted among people, and hence, the use of regression models like support vector regression and multi-view adaptive regression [7] was quite prevalent in the initial years of study. Deep learning models came into the picture later on but in recent times they are being widely used for this task ([4],[7],[16],[20],[23]-[26]). Modifying the memorability of images was initially done by classic photo editing software [8]. Later, Generative Adversarial Networks (GANs) [9] gained popularity for modifying image memorability ([28]-[30]). Generative models ([28],[29]) find various applications in creating images from text, modifying images, creating images based on a given category, and so on.

Image memorability has many applications [6], such as in education, where we try to create more memorable academic materials to help students memorize better. In this research paper, we present a comprehensive literature survey on various deep learning methods used for the prediction and modification of the visual memorability of images. In particular, we analyze the use of Convolutional Neural Networks, Recurrent Neural Networks, and Generative Adversarial Networks for image memorability prediction and modification. Our findings will aid others in understanding and researching recent trends in predicting and modifying image visual memorability using deep learning techniques. The following describes how the paper was put together. Section 2 describes the dataset used to compare various works. After this, section 3 describes the various deep learning techniques used in the prediction of memorability while section 4 discusses the deep learning techniques used in the modification of memorability. Section 5 discusses some general limitations noticed in the works reviewed for this survey and Section provides a summary of the survey and the future scope.

## 2    Dataset Details

Most works in predicting and modifying image memorability have made use of the LaMem dataset [4]. The authors created this dataset by introducing an optimized protocol of [10] memory games to find out the true memorability scores. In this game, the author shows the images in sequential order. A few of the images shown during the game were repeated. When observers encounter an image they have already seen, they are told to press a button. This experiment helped them to collect real-world data on how memorable images are. It is an enormous dataset containing 60,000 images with their memorability scores. Due to the large size of this dataset, it can be used for training deep neural networks.

## 3    Prediction of Image Memorability with Deep Learning

### 3.1  Convolutional Neural Networks

Convolutional neural networks (CNNs) are deep learning models that can take images as input and perform linear and non-linear operations on them to map them onto the desired output type based on learnable weights and biases. Because CNN immediately learns the features, they eliminate the requirement for human feature extraction.

CNNs are particularly useful for discovering patterns in images to identify scenes, objects, and faces. As the popularity and success of CNNs have increased in computer vision in recent times, they have also been heavily implemented for memorability prediction. Some works have used these models to extract features, while others have proposed an end-to-end deep learning framework based on CNN regressions.

The very first attempt at using CNNs for the task of image memorability prediction was made by Khosla et al. [4] in 2015, where the authors introduced a deep neural network based on AlexNet [11] to extract features of an image which were passed through three fully connected layers and an additional Euclidean loss layer was added since memorability is a single real-valued output. The final prediction was made using SVR. The AlexNet model used in the proposed architecture, MemNet, was initially trained on two popular datasets commonly used for image classification, ILSVRC 2012 [12] and the Places dataset [13]. The final MemNet model was then trained on the LaMem dataset, the largest fully-annotated image memorability dataset available [4]. The authors reported MemNet having a Spearman's rank correlation of 0.64 on training and testing with the LaMem dataset. Although this work opened doors for the wide use of deep learning in image memorability prediction, this approach had a few drawbacks that did not age well with time. Many researchers found difficulty in reproducing the results obtained from the original MemNet model because the original model was implemented on the Caffe framework, which has now been discarded. Moreover, it was observed by researchers like Needell et al. [14] when generalizing the MemNet model on a new dataset, its performance is reduced.

The subsequent major work in this domain was done by Baveye et al. in 2016 [7]. The authors used transfer learning and used models trained to predict object and scene semantics by fine-tuning them for predicting memorability. They chose GoogleNet CNN as their baseline model [15]. They have fine-tuned this model by replacing the two auxiliary losses in the intermediate layers and the SoftMax activation in the final layer with one final fully-connected layer. Their model, MemoNet, achieved a Spearman's rank correlation of 0.636 when trained with 30K training iterations. Unlike in MemNet [4], MemoNet was tested on a mixed dataset and ensured a balance in the emotional feature distribution of the dataset's images. The model performed poorly on this testing dataset. The authors stated the reason for this as being that the negatively aroused images tend to have a more predictable memorability as compared to neutral or positively aroused ones.

In 2018, Squalli-Houssaini et al. [16] took a different approach than most of their peers at the time for predicting images' memorability. They considered not only the visual features but also the semantic features obtained from an image and its textual representation. For the extraction of visual features, the authors used a VGG16 CNN model [17] that had been pre-trained on the ImageNet dataset [18]. As for the extraction of semantic features, they used an Image Captioning (IC) system obtained from the model proposed by Kiros et al. [19]. This system consisted of a CNN and an LSTM network to encode simultaneous image-text embeddings. Using these visual and semantic features, they trained an SVR model for regression and a Multi-Layer Perceptron for classification. To classify the image in the order of its memorability, they converted the memorability scores in the LaMem dataset into class labels. The regression model with SVR resulted in a Spearman's rank correlation at par with MemNet (0.64), while the classification approach resulted in a performance improvement. One major drawback of this work is that this model does not generalize well as it yields a poor performance when tested on some other datasets.

Perera et al. [20], in their work on memorability prediction, also took a similar approach in using transfer learning. However, their work differed as they noted that instead of fine-tuning an entire pre-trained CNN model, only fine-tuning the final layer of a CNN model yielded an improvement in overall model performance. This performance enhancement was suggested to be due to the likely overfitting of the models in previous works. Their model, MemBoost, follows a base setup of MemNet and implements it on different CNN models, regression algorithms, and datasets. The best performance was achieved when a ResNet-152 [21] was pre-trained on a combination of ImageNet [18] and Places datasets [13]. An ensemble regression algorithm, XGBoost [22], was used to predict the final memorability scores. This approach achieved a Spearman's rank correlation of 0.67 on the LaMem dataset.

A new approach using multiple instance learning-based CNNs was proposed by Basavaraju et al. [23] in 2019. Their proposed model, EMNet, considers the features corresponding to various emotions induced, as some emotions are correlated with a higher likelihood of being remembered. This model takes an ensemble of two deep CNNs. The first part of the model consists of a VGG-16 [17] to extract object features and predict their memorability scores. The second part of the model extracts the emotional features. This is executed by a novel framework that uses emotion and salience features for memorability prediction. Using the combined memorability scores from these two models, the final output is produced. As a result of this ensemble deep learning framework, a Spearman's rank correlation of 0.671 is achieved.

In 2020, Zhu et al. proposed a multi-task deep learning approach using aesthetic attributes [26]. The authors believed there to be a hidden relationship between the aesthetics of an image and its memorability. Thus, they proposed a framework that they alternatively trained on an aesthetics dataset (AADB dataset [27]) and a memorability dataset (LaMem [4]). In this approach, they also used a pixel-wise visual attention mechanism (PiCANet). They have used this model to generate attention maps at each pixel and embedded it in a CNN architecture. The authors report that their proposed approach obtained a Spearman's rank correlation of 0.67. This is the first attempt associating aesthetics assessment with memorability to predict memorability. The enhanced performance indicates that the two are correlated and that an analysis of aesthetics can be used to assist in the prediction of memorability.

### 3.2 Recurrent Neural Networks

Recurrent Neural Networks are a particular type of neural network that takes as sequential input data, and at each step, the output from its previous layers is passed as input for the current step. RNNs are capable of remembering the sequence of data.

This property gives these networks a wide range of applications in natural language processing, speech recognition, and time series forecasting. Long Short-Term Memory is one of the most common types of RNNs. As a result of their ability to capture long-term dependencies in data, LSTMs are highly popular. In recent times, the task of predicting the visual memorability of images is greatly benefiting from the use of LSTMs.

Fajtl et al. [25], in their work on predicting the memorability of images, propose an end-to-end deep learning framework. Their proposed attention-based model, AMNet, can be divided into four elements - a deep CNN pre-trained on the ImageNet dataset; a soft attention mechanism to aid the network selectively focus on certain areas of an image; an LSTM for estimating the memorability of the image; lastly, a fully connected layer to output the memorability regression scores. This framework first extracts features of an image using ResNet-50 [21], then these features are passed through a soft attention layer that generates attention maps to highlight regions to focus more on. These maps are then passed to LSTM and a fully connected layer to obtain the prediction results. The authors report their proposed approach to have obtained a Spearman's rank correlation of 0.67 on the LaMem dataset.

A more recent work in memorability prediction using RNN is ResMem-Net, a model introduced by Praveen et al. [24]. Unlike most of the previous work, ResMem-Net does not just use the activations of the final layer to make the prediction. The framework is such that there is a pre-trained ResNet-50 [21] at the top being used as the baseline model. Global Average Pooling is used to pass the hidden layers of ResNet-50 to an LSTM recurrent unit. The output from the LSTM is passed through a fully connected layer to obtain the memorability score. LSTM units can retain all the vital information obtained from activating the hidden layers of the ResNet-50 model, which makes the framework robust. The authors report that their proposed architecture achieved a Spearman's rank correlation of 0.679, making ResMem-Net the current state-of-the-art model.

### 3.3 Comparing the performance of different prediction models

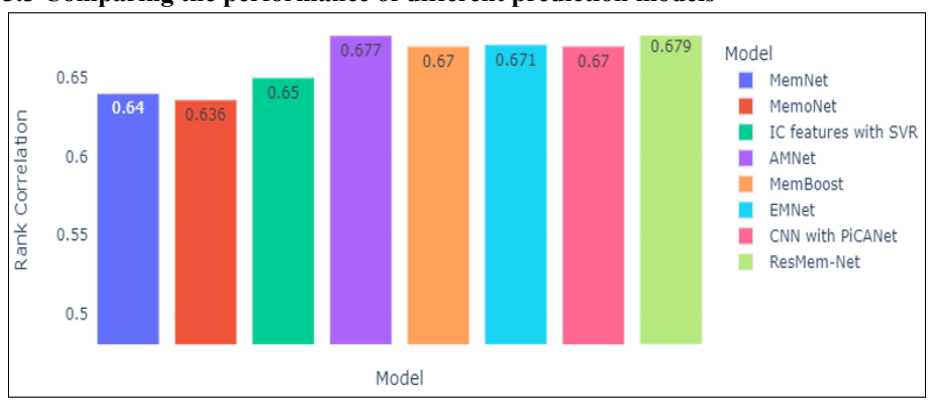

**Figure 1.** Bar chart comparison for different prediction models based on their Spearman's Rank Correlation between the predicted values and ground truth

**Table 1:** Comparison of different prediction models based on their Spearman's Rank Correlation between the predicted values and ground truth

| Method (on LaMem dataset) | Spearman's Rank Correlation (ρ) |
|---|---|
| MemNet (Khosla et. al [4]) | 0.64 |
| MemoNet (Baveye et. al [7]) | 0.636 |
| IC features with SVR (Squalli-Houssaini et. al [16]) | 0.65 |
| AMNet (Fajtl et. al [25]) | 0.677 |
| MemBoost (Perera et. al [20]) | 0.67 |
| EMNet (Basavaraju et. al [23]) | 0.671 |
| CNN with PiCANet (Zhu et. al [26]) | 0.67 |
| ResMem-Net (Praveen et. al [24]) | 0.679 |

Table 1 demonstrates Spearman's rank correlation values of the discussed memorability prediction models that use deep learning. It can be seen that ResMem-Net outperforms the rest of the prediction models. Figure 1 helps visualize the comparison between these models. A general trend can be noted that attention-based neural networks tend to perform better on the task of memorability prediction. This could be attributed to the visual attention mechanism's ability to focus on parts of the picture that provide more relevant information. The memorability of images is often determined by the memorability of the objects present in them, and thus, an attention mechanism can identify these objects and give an enhanced performance. Another trend that can be observed is that LSTMs, in general, give a better performance in this task. Only knowing the parts of the image to focus on is not enough for the task of memorability. The correlations between parts of the image must also be considered. This requires a memory unit to be aware of the entire image at a time. That is why LSTMs are becoming popular in memorability prediction, as they can model long-range dependencies.

## 4 Modification of Image Memorability with Deep Learning

### 4.1 Generative Adversarial Networks

Generative Adversarial Networks (GAN) are a class of deep learning models that can be used to generate a wide variety of natural-looking pictures with subtle variations in their aesthetic characteristics. GANs are made up of two parts: a generator that tries to create actual data and a discriminator that seeks to distinguish between actual and generated data. GANs have been used for picture translation tasks as well as to create photorealistic pictures. Previously efforts have been made to extract the "memorability" component from already trained GANs, enabling the creation of pictures that are thought to be memorable. Such a method still relies on repetitive detection models and calls for a seed picture whose memorability is later modified.

Sidorov et al. [28] demonstrate how cutting-edge deep learning methods created for regulated image-to-image translation can enhance visual memorability. The actual media creators—photographers, designers, and other artists—cannot employ advanced mathematical or behavioral features that may affect image memorability in their work. Hence this work examines how commonly used photo editing tools that people use regularly affect visual memorability. This experiment used the data set to train algorithms like VAE/GAN, StarGAN, and AttGAN. However, only AttGAN produced a fruitful outcome. This method resulted in changes of up to 33% and operated effectively in the two directions (increasing and decreasing memorability). Further, the study demonstrated that essential image processing technologies could not cause changes in the memorability of an image expectedly. The analysis of the data showed that procedures like blurring, darkening, and discoloration that cause information loss also result in a reduction in memorability. Sharpening was the only method that consistently improved the memorability score when applied to an image.

In 2019, Goetschalckx et al. [29] proposed a new model using GANs to study the memorability of an image. The model architecture consisted of a transformer, an assessor, and a generator. The input taken for this model was the latent space of images, not the actual images. A set of 4,00,000 latent space vectors was used to implement this model. The GAN model used here was the BigGAN model pre-trained on the ImageNet [18] dataset, while the assessor model implemented was the CNN MemNet. This study provided a methodology for visualizing what a GAN-based model has discovered about its target picture attribute from another model (the assessor in this case). This study can be expanded by substituting the Assessor model in the present framework for any other image properties that are difficult to comprehend and visualize, for example, the emotional valence of an image. Through behavioral human memory testing using edited photos, it was confirmed that the model successfully changed GAN images to make them more (or less) memorable.

The first example of a GAN trained from the start specifically to create memorable landscape images was presented by Davidson et al. [30]. It made use of information on two-dimensional memorability gathered from human experiments. To produce eye-catching graphics, a powerful generative model, such as a Wasserstein GAN (WGAN), is adjusted. The study evaluated the model's output and looked at how memorability levels affected how "genuine" the visuals were that were produced. Using an independent memorability prediction network, the authors discovered that images designed to be recalled were more memorable than images designed to be less memorable. Therefore, the formation of semantic features and the spatial relationships between these elements are controlled by the memorability of a picture.

The trends in recent studies on image memorability modification demonstrated the use of GANs for generating natural-looking pictures with subtle variations in their image characteristics. Image memorability is one of the characteristics that can be modified to produce a continuum of images. The studies demonstrated that image characteristics like blurring, darkening, sharpening, etc. affect the memorability of the images in different ways. The visualizations in the various previous studies presented several potential elements that might contribute to the understanding of why some images are more memorable than others. It was also found that memorability may sometimes affect the level of realness a picture holds.

## 5      Limitations

In the earliest works on predictability, the main issue faced by researchers was to find and keep the memorable elements of a picture while enhancing forgettable ones. They aimed at finding the parts of images that are memorable or forgettable so that information consumption can be made easier. Later it was found that an emotional bias also influences the memorability of any image. This proved memorability to be more subjective and dependent on the observer. If one is simply interested in the fundamental information of the pictures, the memorability forecast will be inaccurate. After some work provided sufficient results on the Lamem dataset, working on a larger dataset was the next challenge involved. Some factors were considered for the dataset creation. These factors included the number of observers, the validity of preserving one score per image, the necessity of reconsidering the sequence of the photos, etc. A larger dataset is said to perform better with respect to the predictions. This is because a large dataset provides ease in generalizing neural networks. The ResMemNet model is one of the latest CNN-based models for image memorability prediction. The main challenges that this model faced were that it was not deployable on any Mobile GPU hence constraining its viability and applications. Furthermore, the ResNet model and LSTM units were not the most latest and efficient in this study.
The research in image memorability modification majorly faced challenges in correlating the psychophysical features of images with their memorability scores. Researchers failed to establish a more general way of modifying the images by increasing or decreasing the image memorability by providing an image input to the model. The current models lack an image-to-image translation method of memorability score modification.

# 6    Conclusion and Future Scope

By performing a comparative study on the previous works in image memorability prediction, it can be established that ResMem-Net outperforms the rest of the prediction models. It is generally accepted that attention-based neural networks outperform other types of neural networks when it comes to predicting memorability. Another pattern that may be seen is that LSTMs do better in this job overall. The challenge of memorability requires more than simply understanding which portions of the image to concentrate on. It's also important to take into account how the image's components correlate. The use of GANs for creating natural-looking images with small variations in their visual attributes was shown by trends in recent studies on image memorability modification.  The results showed that numerous picture qualities, such as blurring, darkening, and sharpening, had an impact on how memorable an image is. The visualizations used in the earlier research provide many potential components that might help explain why certain pictures are more remembered than others. Additionally, it was discovered that occasionally a picture's memorability might influence how realistic it seems.

Various works on image memorability prediction and modification using deep learning were reviewed in this survey, and some trends were noticed. To summarize, predicting image memorability has dramatically benefited from deep learning and is achieving near human-level consistency. However, there are some questions on the validity of current memorability scores, with only one significant large-scale image memorability dataset available. The progress of modifying image memorability is relatively slow-paced, even with the introduction of deep learning in this domain. Many challenges need to be addressed to modify the memorability of images successfully. Nevertheless, image-to-image translation GANs are showing promising results.

In the quest of using deep learning techniques to accurately predict and modify the memorability of images, despite its success, there are still areas where further research is required. Deep learning is heavily dependent on the size and diversity of data available. There is a need to create more large-scale datasets because there is currently just one substantial large-scale image memorability dataset available. In memorability prediction, it was noted that other cognitive attributes such as emotional bias also impact the memorability of images. A future direction could be to further explore the relationship between different cognitive attributes with memorability and take them into consideration while training the deep learning models. In memorability modification, GANs have shown impressive results and a further direction in this could be to carefully condition the network. Moreover, methods like stable diffusion can also be experimented with to yield more realistic images.